\documentclass[11pt]{article}

\usepackage[preprint]{acl}

\usepackage{times}
\usepackage{latexsym}
\usepackage[T1]{fontenc}
\usepackage[utf8]{inputenc}
\usepackage{microtype}

\usepackage{amsmath}
\usepackage{hyperref}
\usepackage{url}
\usepackage{booktabs}
\usepackage{enumitem}
\usepackage{amsfonts}
\usepackage{nicefrac}
\usepackage{graphicx}
\usepackage{multirow}
\usepackage{siunitx}
\usepackage{algorithm}
\usepackage{float}
\usepackage{needspace}
\usepackage{placeins}
\usepackage{algpseudocode}
\usepackage{pifont}
\usepackage{amssymb}
\usepackage{makecell}
\usepackage[table]{xcolor}

\title{Write, Execute, Refine: From Skill Follower to Skill Optimizer\\ via Reinforcement Learning from Execution Feedback}

\author{
  \textbf{Kang Peng}\textsuperscript{1}\thanks{Equal contribution.},
  \textbf{Zhiwei Zhang}\textsuperscript{2,3}\footnotemark[1],
  \textbf{Yichen Zhang}\textsuperscript{4},
  \textbf{Zezhong Wang}\textsuperscript{2},
  \\
  \textbf{Yiming Du}\textsuperscript{2},
  \textbf{Geng Tu}\textsuperscript{1},
  \textbf{Baojun Wang}\textsuperscript{5},
  \textbf{Bin Liang}\textsuperscript{2,3},
  \textbf{Ruifeng Xu}\textsuperscript{1}\thanks{Corresponding author.},
  \textbf{Kam-Fai Wong}\textsuperscript{2,3}
  \\[2mm]
  \textsuperscript{1}Harbin Institute of Technology, Shenzhen, China
  \\
  \textsuperscript{2}The Chinese University of Hong Kong
  \\
  \textsuperscript{3}MoE Key Laboratory of High Confidence Software Technologies
  \\
  \textsuperscript{4}Harbin Institute of Technology, Harbin, China
  \\
  \textsuperscript{5}Huawei Technologies Co., Ltd.
  \\[1mm]
  \texttt{26s165138@stu.hit.edu.cn, zhangzhiwei1019@link.cuhk.edu.hk}
}

\begin{document}
\maketitle

\begin{abstract}
Expert-written natural language skills can improve tool-using agents, yet agent-authored skills perform 8--11 points worse than using no skill.
While inference-time loops can repair individual skills per task, they leave the underlying generator's refinement capability unimproved.
We study how to organize execution experience from intermediate skills into training states for an optimizer.
We introduce \textbf{WER} (\textbf{W}rite, \textbf{E}xecute, and \textbf{R}efine), a multi-phase framework that trains a Skill Optimizer outside a frozen executor.
The optimizer proposes skills, a frozen agent executes each repeatedly, and a programmatic verifier scores the outcomes.
These scores serve as group-relative rewards and filter for informative, mixed-outcome instances.
Matched successful and failed trajectories from these records form the next phase's refinement states, so the optimizer learns from the consequences of its earlier outputs.
On BFCL v4 multi-turn and $\tau^{2}$-bench, WER improves average Pass@1 over the no-skill baseline by 7.80 and 6.06 points, respectively.
Under an identical refinement workflow, it outperforms the same backbone without optimizer training by 10.02 and 12.50 points.
Notably, our trained 4B optimizer achieves 76.63\% on BFCL v4, outperforming all evaluated off-the-shelf general-purpose LLMs in the same role on average. Our code is available at \url{https://github.com/littlepkk/wer-training}.

\end{abstract}

\section{Introduction}
\label{sec:intro}

Reliable tool use requires procedural knowledge: which tool to invoke, how to validate its arguments, and when to retry~\cite{yao2022react,schick2023toolformer,patil2025berkeley,zhang2026robust}.
A common approach represents this knowledge as \emph{skills}, compact natural-language instructions added to an agent's context at inference time~\cite{wang2023voyager,jiang2026sok}.
On SkillsBench~\cite{li2026skillsbench}, expert-curated skills raise the average pass rate from 33.9\% to 50.5\%, yet agent-authored skills fall 8--11 points below using no skill at all.
This reveals a capability gap: agents can benefit substantially from procedural guidance without being able to write it reliably.

\begin{figure}[t]
    \centering
    \includegraphics[width=\columnwidth]{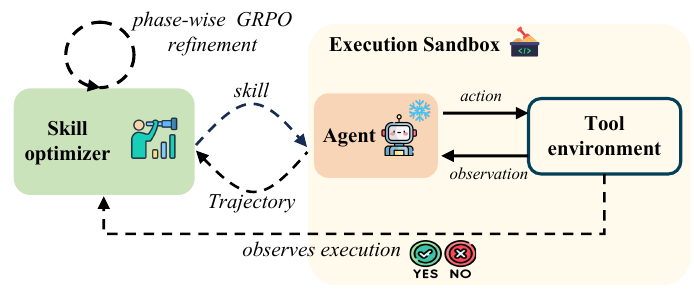}
    \caption{%
    \textbf{The Skill Optimizer remains outside the environment and observes execution.}
    It writes a skill for a \emph{frozen} agent and revises the skill from the resulting trajectory.%
    }
    \label{fig:teaser}
\end{figure}

A common response to this gap is an inference-time loop in which an LLM drafts a skill, observes its execution, and revises it~\cite{shinn2023reflexion,liu2025contextual,alzubi2026evoskill,yang2026skillopt,liu2026skillrevise}.
More broadly, reflecting on trajectories and execution outcomes, then distilling the lessons into reusable experience, has become a common workflow recipe for improving agents.
Yet this recipe presupposes a capability that language models may not reliably possess: converting behavioral evidence into the procedural instruction needed to prevent the observed failure.
While inference-time refinement can improve individual skills, it does not directly train the model to perform better execution-grounded reflection.
Consequently, refining skills for subsequent tasks still relies on the model's existing, uncalibrated ability to diagnose failures and formulate actionable corrections.
Worse still, without grounded reflection capabilities, models frequently hallucinate plausible yet erroneous guidance.
For instance, injecting merely 10\% flawed experience reduces $\tau^{2}$-bench Pass@1 from 82.5 to 77.2, an error that self-verification fails to recover~\cite{zhu2026escaping}.

Recent reinforcement-learning methods have established that skill improvement can itself be parameterized and learned.
SkillMaster~\cite{yang2026skillmaster} discovers post-episode mutations through counterfactual probes; SkillOS~\cite{ouyang2026skillos} trains repository management based on the long-term utility of updates; and Skill-R1~\cite{vishe2026skillr1} optimizes recurrent skill generation via intra- and inter-generation advantages.
Together, these works shift the paradigm from prompting fixed LLMs to optimizing a specialized refinement policy.
Within this emerging direction, we study a central question: \emph{how should execution experience from intermediate skills be structured into progressive training states for a skill optimizer?}
Rather than treating refinement as isolated steps, we argue that effective training requires exposing the optimizer to the dynamic consequences of its own prior revisions across successive iterations.

We instantiate this idea as \textbf{WER} (\textbf{W}rite, \textbf{E}xecute, and \textbf{R}efine).
As shown in Figure~\ref{fig:teaser}, a dedicated Skill Optimizer $\pi_\theta$ operates outside the execution sandbox.
Given a refinement state, $\pi_\theta$ proposes candidate skills; a frozen agent executes each candidate repeatedly; and a deterministic verifier provides group-relative rewards.
To construct informative supervision across phases, WER identifies candidates with mixed outcomes (where the same skill yields both successful and failed runs under identical environments).
Pairing these contrasting trajectories provides an isolated, controlled comparison of divergent behavioral paths.
WER reassembles this paired evidence with the corresponding intermediate skill into the refinement state for the next phase, enabling the optimizer to self-bootstrap from its own developmental trajectory without modifying downstream executor weights.

\Needspace{4\baselineskip}
Our main contributions are:
\begin{itemize}[leftmargin=*,nosep]
    \item We formulate skill refinement as an autonomous, execution-grounded policy learning problem, decoupling the Skill Optimizer from a frozen downstream agent to optimize natural-language guidance directly from multi-turn interaction traces.
    \item We introduce \emph{phase-wise self-bootstrapping}, a multi-phase framework that pairs matched successful and failed executions from the optimizer's own prior outputs into contrastive diagnostic states, enabling the policy to learn progressively from its earlier revisions.
    \Needspace{11\baselineskip}
    \item Under an identical refinement workflow, WER improves average Pass@1 over the untrained backbone by 10.02 points on BFCL v4 multi-turn~\cite{patil2025berkeley} and 12.50 points on $\tau^{2}$-bench~\cite{barres2025tau}. Remarkably, our 4B optimizer reaches 76.63\% on BFCL v4, outperforming all evaluated off-the-shelf frontier LLMs deployed in the same optimizer role.
\end{itemize}

\section{Related Work}
\label{sec:related}

\paragraph{Skill Construction and Inference-Time Refinement.}
External skill artifacts provide modular procedural knowledge that can be created, retrieved, and revised without modifying model weights.
Early approaches construct persistent skill libraries from sandbox exploration~\cite{wang2023voyager,ni2026trace2skill}, while recent benchmarks systematize their lifecycle and utility in tool use~\cite{li2026skillsbench,jiang2026sok}.
To correct execution errors, a prominent line of work employs test-time reflection loops, iteratively editing skill documents via failure analysis, execution constraints, or multi-agent verification~\cite{alzubi2026evoskill,yang2026skillopt,liu2026skillrevise,zhu2026escaping}.
At a broader level, this parallels prompt-optimization frameworks that search over textual artifacts using feedback or textual gradients~\cite{yang2024opro,wang2024promptagent,guo2024connecting,khattab2024dspy,yuksekgonul2024textgrad}.
However, these inference-time strategies treat the skill author as a fixed, frozen black box.
Consequently, each new task requires fresh trial-and-error reasoning without accumulating systematic refinement expertise into the authoring model itself.

\paragraph{Skill Internalization and Executor Co-Training.}
Another line of research integrates skills directly into agent training loops.
Several methods co-train skill selection and utilization alongside downstream execution policies~\cite{wang2026reinforcement,shi2026skill1,xia2026skillrl,he2026reskill}.
Others focus on parameter internalization, progressively pruning or absorbing external skills into the executor's parametric memory through continued reinforcement learning~\cite{lu2026skill0,zhu2026skill0.5}.
While effective at specializing the acting agent, these approaches fundamentally couple skill evolution with executor parameter updates, making them computationally prohibitive when deployed across distinct or proprietary black-box execution engines.
In contrast, WER deliberately freezes the downstream executor, isolating skill refinement as an autonomous, transferable meta-policy over natural language prompts.

\paragraph{Learned Skill Management and Meta-Optimization.}
Closest to our work are methods that parameterize skill-level decision-making.
SkillMaster~\cite{yang2026skillmaster} evaluates post-episode skill mutations counterfactually against probe tasks; SkillOS~\cite{ouyang2026skillos} optimizes library curation based on downstream lifelong utility; and Skill-R1~\cite{vishe2026skillr1} trains a lightweight skill generator using recurrent intra- and inter-generation rollout advantages.
WER differs in how execution experience is synthesized into supervision.
While prior learned managers optimize updates against scalar probe rewards or sequential task streams, WER introduces phase-wise self-bootstrapping: it isolates mixed-outcome episodes from the optimizer's own prior outputs and pairs matched successful and failed trajectories into contrastive refinement states.
This provides the optimizer with localized behavioral counter-evidence, grounding policy updates directly in fine-grained procedural discrepancies rather than unconditioned outcome signals.

\section{Method}
\label{sec:method}

\begin{figure*}[t]
    \centering
    \includegraphics[width=\textwidth]{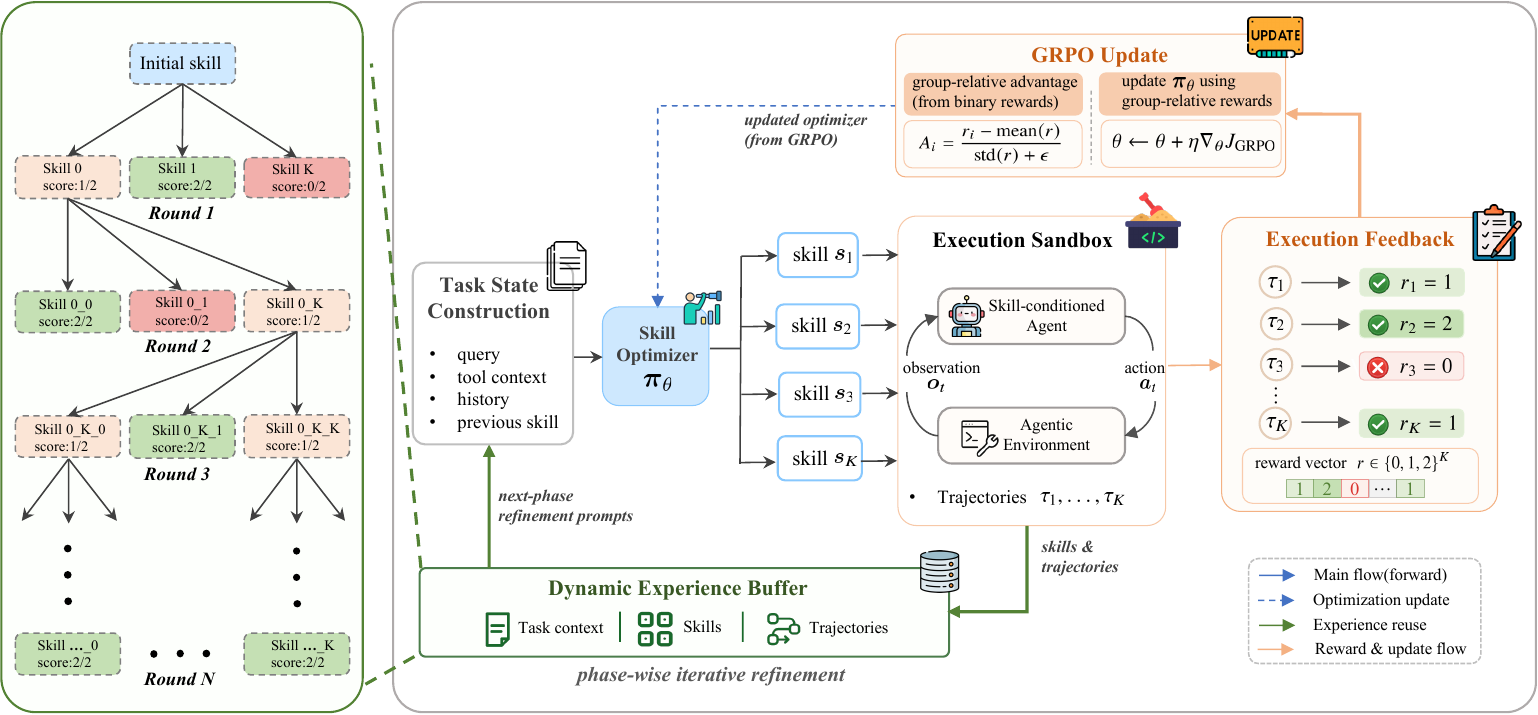}
    \caption{%
    \textbf{Training loop.}
    A refinement state is assembled from the task context, interaction history, and current skill, from which the Skill Optimizer samples $K$ candidate skills (\S\ref{sec:operator}).
    A frozen skill-conditioned agent executes each candidate in the sandbox, producing trajectories and verifier outcomes (\S\ref{sec:observe}).
    The outcomes provide group-relative advantages for updating $\pi_\theta$ (\S\ref{sec:grpo}), while the skills and trajectories enter the experience buffer to assemble next-phase states (\S\ref{sec:phase}).
    As illustrated on the left, retained skills seed the candidates of the next round, forming a revision tree across phases.
    For clarity, the figure shows binary verifier rewards; the actual reward also includes the format and length terms in Eq.~\ref{eq:reward}.%
    }
    \label{fig:framework}
\end{figure*}

We consider an agent that solves multi-turn tasks by calling tools in a stateful environment, guided by a short natural-language skill.
An episode succeeds when its outcome matches the reference solution.
We train a policy to revise the current skill from its execution record, with the executor frozen.
Figure~\ref{fig:framework} shows the loop.
We instantiate it in two multi-turn, stateful tool environments that expose programmatic verifiers, BFCL multi-turn~\cite{patil2025berkeley} and $\tau^{2}$-bench~\cite{barres2025tau}, and defer the full configuration to the experimental setup.
We define the operator (\S\ref{sec:operator}), its execution feedback (\S\ref{sec:observe}), group-relative optimization (\S\ref{sec:grpo}), and phase-wise self-bootstrapping (\S\ref{sec:phase}).

\subsection{Skill Refinement as a Learned Operator}
\label{sec:operator}

\paragraph{Skills.}
A skill is a Markdown document with four sections: a name, a one-line task-family description, a numbered workflow, and notes on assumptions, edge cases, and known failure modes.
It specifies a reusable procedure without instance-specific argument values or entity identifiers; its body is prepended to the executor's system prompt.

\paragraph{The operator.}
Let
\begin{equation}
    x = (q,\; \mathcal{C},\; h,\; s)
    \label{eq:state}
\end{equation}
denote a \emph{refinement state}, where $q$ is the user query, $\mathcal{C}$ the tool definitions and environment context, $s$ the skill currently in force, and $h$ the interaction history consisting of trajectories collected by executing $s$ on the task, annotated by their verified outcomes.
The Skill Optimizer is a policy over skill documents,
\begin{equation}
    s' \sim \pi_\theta(\,\cdot \mid x\,).
    \label{eq:operator}
\end{equation}
The optimizer outputs only skill documents and influences execution through the injected skill.


All refinement rounds use Eq.~\ref{eq:operator}.
Initially, $s$ is an external seed skill (generated by GPT-5.1) and $h$ contains its execution trajectories.
Later rounds pass the revised skill and its updated history back to the same operator.
We refer to one operator application as a \emph{round}, and a dataset-wide training stage as a \emph{phase}.
Retained records from each phase are pooled to construct successor states, so the number of phases bounds the training revision depth.
The number of inference-time rounds can be chosen independently.

\subsection{What the Optimizer Observes}
\label{sec:observe}

The optimizer receives execution trajectories as context and verified outcomes as rewards.

\paragraph{Execution.}
A candidate skill $s'$ is injected into a frozen agent $\pi_A$, which interacts with an instrumented environment $\mathcal{E}$ and produces a trajectory
\begin{equation}
    \tau \;=\; \big(a_1, o_1, \ldots, a_T, o_T,\; y\big),
    \label{eq:traj}
\end{equation}
where $a_t$ is a tool call together with its arguments, $o_t$ is the environment response including any error returned, and $y$ is the terminal environment state.
The parameters of $\pi_A$ remain fixed throughout training.

\paragraph{Channel one: the trajectory, as context.}
The trajectory is passed to the optimizer without summarization.
The next refinement state preserves the call sequence verbatim, including which tool was chosen, its arguments, the environment response, and the agent's actions after a failure.
A scalar reward reports whether execution succeeded but does not identify the procedural guidance that needs revision.
For example, a zero reward alone does not explain the failure.
If the trajectory shows the same call issued three times against an identifier the environment has already invalidated, it suggests that the skill needs to instruct the agent to check state before acting.

\paragraph{Channel two: the outcome, as reward.}
The environment verifier determines whether $s'$ succeeds.
In BFCL multi-turn we compare the final state of every environment object against the state reached by the reference solution; in $\tau^2$-bench we compare the terminal database against the reference database and check that the required actions were taken.
Both checks are programmatic and deterministic.
As discussed in \S\ref{sec:intro}, using the same model family to generate behaviour and score it can compromise the quality signal.
We combine a frozen executor with programmatic verification to isolate the effect of skill revisions and obtain scores independent of model judgments.

\paragraph{Attribution, and its limits.}
With $\pi_A$ fixed, candidates for the same $x$ are evaluated under a common executor, allowing outcome differences to be attributed to their skill documents in expectation.
Individual trajectories also reflect executor sampling, so we aggregate $n$ rollouts per candidate.
We use a small optimizer to keep training tractable and a larger executor for its task-solving capability.

\subsection{Group-Relative Optimization}
\label{sec:grpo}

Absolute outcome varies far more across tasks than across the candidate skills written for any one of them, so an absolute reward would mostly measure task difficulty.
We therefore score candidates against each other under a matched refinement state.

\paragraph{Reward.}
For a refinement state $x$ the optimizer samples $K$ candidates, each evaluated over $n=2$ rollouts.
Let $o$ denote the full completion, comprising a reasoning block and the extracted skill document $s'$.
The scalar reward at global training step $t$ combines three terms,
\begin{equation}
    R(o; t) \;=\; \tfrac{1}{3}\Big( \alpha_t I_{\mathrm{fmt}}(o) + R_{\mathrm{task}}(s') + R_{\mathrm{len}}(o; t) \Big),
    \label{eq:reward}
\end{equation}
with the following roles.
$I_{\mathrm{fmt}}(o) \in \{0,1\}$ checks format compliance; its weight $\alpha_t$ anneals downward to shift optimization focus toward task completion.
$R_{\mathrm{task}}(s') = \sum_{j=1}^{n} r_{\mathrm{ver}}(\tau_j) \in \{0,1,2\}$ sums binary verifier outcomes across rollouts.
$R_{\mathrm{len}}(o; t) \in [0,1]$ constrains the reasoning block, penalising empty traces and length overruns with a dynamic budget (details in Appendix~\ref{app:training-details}).

\paragraph{Why the task reward stays coarse.}
Finer-grained rewards could assess tool selection, argument correctness, error recovery, and confirmation before irreversible actions.
These judgments exceed the available programmatic checks and would require model-based scoring, introducing the concern discussed in \S\ref{sec:intro}.
We use the verified task outcome and compare candidates within each group.
Two candidates for the same state differ only in their text, so a consistent gap between them is informative even when the individual signal is coarse.

\paragraph{Update.}
The advantage of a candidate is its deviation from the group mean,
\begin{equation}
    \hat{A}_k \;=\; R_k - \mu(\mathbf{R}), \qquad \mathbf{R} = (R_1,\ldots,R_K),
    \label{eq:adv}
\end{equation}
optionally rescaled by $\sigma(\mathbf{R})$.
The optimizer is updated with the clipped GRPO surrogate~\cite{shao2024deepseekmath},
\begin{equation}
    \resizebox{\columnwidth}{!}{$\displaystyle
    \mathcal{J}(\theta) = \mathbb{E}_{x}\!\left[\frac{1}{K}\sum_{k=1}^{K}\min\!\Big(\rho_k \hat{A}_k,\ \mathrm{clip}(\rho_k, 1{-}\epsilon, 1{+}\epsilon)\,\hat{A}_k\Big)\right]
    $}
    \label{eq:grpo}
\end{equation}
where $\rho_k = \pi_\theta(o_k \mid x)/\pi_{\text{old}}(o_k \mid x)$ denotes the probability ratio evaluated over the complete generated completion $o_k$.
We use no KL term and no reference policy.
The output we want is a document format and a revision style that the base model does not yet have, so anchoring the policy to its initialization works against the objective.

\subsection{Phase-Wise Self-Bootstrapping}
\label{sec:phase}

A revision may address a failure exposed by the previous execution, but its effect must be checked by running the revised skill.
This execution can reveal unresolved errors or additional failure modes, providing evidence for the next revision.
Training across phases exposes the optimizer to these successive refinement states, including those produced by its own earlier edits.
Execution in round $t$ thus supplies the training evidence for round $t+1$.
As shown on the left of Figure~\ref{fig:framework}, a retained skill and its execution evidence become the parent state from which the next round's candidates are sampled.
Repeating this buffer--reassembly step extends the revision tree across phases.

\paragraph{The buffer.}
Every scored candidate is written to an experience buffer as a tuple
$\big(x,\; s',\; \{\tau_1,\ldots,\tau_n\},\; R\big)$,
with all $n$ trajectories stored in full.
At phase end, retained buffer records form the next refinement states, so the optimizer continues revising its own previous output.

\paragraph{Retention.}
Not every record is useful as a next-phase input.
A candidate whose rollouts all succeed leaves no failure to diagnose; one whose rollouts all fail leaves no working path to contrast against.
With a binary verifier and $n = 2$ rollouts a candidate scores $0$, $1$ or $2$, and we retain exactly the middle case ($1$).
The kept record therefore holds a \emph{matched pair}: two runs of the same skill on the same task, one of which reached the reference state and one of which did not.
The successful trajectory provides a working path the revised skill should preserve.
The failed trajectory reveals a behavioural branch that may need additional guidance.
With skill, task, and executor fixed across the pair, this comparison helps identify procedural gaps and formulate targeted edits.
All retained records from a phase are pooled to assemble the next phase's refinement states without a per-task quota; tasks without mixed outcomes contribute no successor states.
Retention uses only verifier outcomes, independently of format and length rewards.

\paragraph{Assembling the next refinement state.}
The retained skill becomes the current skill $s$, while the tool context $\mathcal{C}$ and query $q$ are preserved from the original task instance.
The matched execution trajectories are organized into the interaction history $h$: the run that reached the reference state is placed in a success block, and the run that did not in a failure block.
This assembled state $x = (q, \mathcal{C}, h, s)$ is then presented to the optimizer as a single revision request.
The call sequences are preserved verbatim, allowing the optimizer to read the raw trajectories scored by the verifier.



\paragraph{Design Insights.}
Phase-wise self-bootstrapping exhibits several key advantages over static offline datasets.
First, which intermediate skills yield mixed outcomes depends dynamically on the evolving optimizer paired with the fixed executor.
By regenerating and reassembling execution records at each phase boundary, the training distribution remains strictly on-policy, tracking the specific behavioral deficiencies introduced by the optimizer's latest updates.
Second, compared with assigning a single sparse reward at the conclusion of multi-step refinement, evaluating intermediate outputs round by round provides denser, finer-grained credit assignment.
Finally, training and inference depths are decoupled: once trained, $\pi_\theta$ can be applied for an arbitrary number of revision rounds at test time.

The overall training procedure is summarized in Appendix~\ref{app:training-procedure} (Algorithm~\ref{alg:training}).

\section{Experiments}
\label{sec:experiments}

\begin{table*}[!t]
    \centering
    \small
    \setlength{\tabcolsep}{4.2pt}
    \renewcommand{\arraystretch}{1.08}
    \begin{tabular}{lccccc@{\hspace{7pt}}cccc}
        \toprule
        \multirow{2}{*}{Method} & \multicolumn{5}{c}{BFCL v4 Multi-Turn} & \multicolumn{4}{c}{$\tau^2$-bench} \\
        \cmidrule(lr){2-6}\cmidrule(lr){7-10}
        & File System & Trading & Travel & Vehicle & \textsc{Avg.}$\uparrow$
        & Airline & Retail & Telecom & \textsc{Avg.}$\uparrow$ \\
        \midrule
        No Skill
        & 68.42 & 86.84 & 62.16 & 57.89 & 68.83
        & 45.00 & 64.84 & 30.77 & 46.87 \\
        GPT-5.1 Seed Skill
        & 63.16 & 78.94 & 62.16 & 64.86 & 67.28
        & 47.50 & 63.73 & 31.86 & 47.70 \\
        Qwen3-4B as Skill Optimizer
        & 57.89 & 84.21 & 59.46 & 64.86 & 66.61
        & 40.00 & 60.43 & 20.87 & 40.43 \\
        Skill-R1~\cite{vishe2026skillr1}
        & \textbf{71.05} & 84.21 & 64.86 & 67.56 & 71.92
        & 47.50 & 56.04 & 20.87 & 41.47 \\
        Trace2Skill~\cite{ni2026trace2skill}
        & 63.16 & 87.55 & 67.57 & 75.68 & 73.49
        & \textbf{52.50} & 58.34 & 29.67 & 46.84 \\
        SkillOpt~\cite{yang2026skillopt}
        & 68.42 & \textbf{92.11} & 67.57 & 72.97 & 75.27
        & 50.00 & \textbf{68.13} & 38.46 & 52.20 \\
        \midrule
        \rowcolor{blue!8}
        \textbf{WER (Ours)}
        & \textbf{71.05} & 86.84 & \textbf{70.27} & \textbf{78.38} & \textbf{76.63}
        & 50.00 & \textbf{68.13} & \textbf{40.66} & \textbf{52.93} \\
        \bottomrule
    \end{tabular}
    \caption{Main results on BFCL v4 and $\tau^2$-bench (Pass@1, \%). Best results are in \textbf{bold}.}
    \label{tab:main-results}
\end{table*}


\subsection{Experimental Setup}
\label{sec:experimental-setup}

\paragraph{Benchmarks and Metrics.}
\label{par:benchmarks-metrics}
We evaluate WER on two benchmarks for long-horizon tool use: the multi-turn-base tasks from BFCL v4~\cite{patil2025berkeley} and $\tau^2$-bench~\cite{barres2025tau}. We split the 200 BFCL v4 multi-turn-base instances into 50 training and 150 test instances, approximately balanced across four domains. We report domain-level results and an overall average that weights each domain equally. For $\tau^2$-bench, which provides its own subdomain partition, we split the base tasks into training and test sets using the same 1:3 ratio. Because a skill serves as an experience guide distilled by an agent, our primary goal is to assess whether an optimized skill can reliably support successful task completion in a single attempt. We therefore report Pass@1 rather than Pass@$k$, avoiding the extra inference cost of repeated attempts. To reduce variance on these relatively small test sets, we run each evaluation three times and report the average Pass@1. Appendix~\ref{app:benchmark-details} provides further details on both benchmarks.
\paragraph{Data Construction.}
\label{par:data-construction}
Because the optimizer only revises text skills, its action space does not depend on the interface of the task environment. We pool the benchmark-specific training splits defined in \hyperref[par:benchmarks-metrics]{Benchmarks and Metrics}, exposing the optimizer to diverse refinement contexts and encouraging cross-benchmark generalization. GPT-5.1 generates a zero-shot initial skill for each task. These skills serve as the first refinement targets and initiate phase-wise self-bootstrapping, which continually collects and organizes the training data needed for later rounds of refinement. Further details are provided in \S\ref{sec:phase}.
\paragraph{Baselines.}
\label{par:baselines}
We compare WER with six baselines. No Skill lets the base agent solve each task without an external skill. GPT-5.1 Seed Skill gives the agent the initial skill generated by GPT-5.1 without further refinement. Qwen3-4B as Skill Optimizer replaces our trained optimizer with the untrained Qwen3-4B base model~\cite{yang2025qwen3}, while keeping the refinement protocol fixed. This comparison measures the effect of optimizer training. Skill-R1~\cite{vishe2026skillr1} trains a lightweight skill generator with reinforcement learning and uses intra- and inter-generation advantages from verified recurrent rollouts. Trace2Skill~\cite{ni2026trace2skill} extracts lessons from individual trajectories in parallel and organizes them into a transferable skill directory without parameter updates. \mbox{SkillOpt}~\cite{yang2026skillopt} uses execution trajectories to propose bounded skill edits and accepts revisions only when they improve held-out validation performance. We use the same models across methods for stages outside skill optimization, controlling for differences in model capability in these stages and focusing the comparison on optimization. Appendix~\ref{app:baseline-details} provides additional details on Skill-R1, Trace2Skill, and SkillOpt.
\paragraph{Evaluation Protocol.}
\label{par:evaluation-protocol}
We use GPT-4o as the frozen executor for all training rollouts and downstream evaluations.
At inference time, refinement starts from the GPT-5.1 seed skills. The WER-trained optimizer performs two rounds of refinement on the training split, producing a set of skills for each domain. GPT-5.5 then merges them into one domain-specific skill, which is appended to the executor's system prompt and evaluated on the corresponding test domain. To ensure a strictly controlled comparison, all baselines share the same seed skills, with GPT-5.5 used consistently for any skill merging or related auxiliary operations outside skill optimization. We use the same protocol for BFCL v4 and $\tau^2$-bench.
The optimizer is trained with GRPO using the verl framework~\cite{sheng2025hybridflow}. The full training configuration is provided in Appendix~\ref{app:training-details}.
\subsection{Main Results}
\label{sec:main-results}

We first ask whether WER produces skills that outperform skill-authoring baselines under a single-attempt protocol. Table~\ref{tab:main-results} reports Pass@1 on both benchmarks. We examine the results from two angles: the gain a WER skill provides over the base agent and the isolated contribution of optimizer training.

\paragraph{WER Improves Agentic Performance of Base LLMs.}
\label{par:main-performance}
Table~\ref{tab:main-results} shows that WER raises the domain-averaged Pass@1 over the No Skill baseline from 68.83\% to 76.63\% on BFCL v4 and from 46.87\% to 52.93\% on $\tau^2$-bench, gains of 7.80 and 6.06 percentage points. On BFCL v4, WER improves three of the four domains and matches the No Skill baseline in Trading. It obtains the best or tied-best result in File System, Travel, and Vehicle. On $\tau^2$-bench, WER improves all three domains, ties for the best Retail score, and achieves the highest Telecom score. Relative to the fixed GPT-5.1 seed skills, WER gains 9.35 points on BFCL v4 and 5.23 points on $\tau^2$-bench. This margin indicates that the gains stem from effective skill optimization rather than solely from GPT-5.1's ability to generate the initial skills. These gains also remain substantial with alternative merging models, suggesting that WER's improvements do not primarily depend on GPT-5.5's merging capability. A more detailed analysis of this comparison is provided in Appendix~\ref{app:merging-model-analysis}.
\paragraph{WER Learns to Refine Skills from Execution Feedback.}
To isolate the effect of optimizer training, we use the untrained Qwen3-4B base model as the skill optimizer and keep the rest of the refinement protocol fixed (see \hyperref[par:evaluation-protocol]{Evaluation Protocol}). It underperforms the GPT-5.1 seed skills on BFCL v4 (66.61\% versus 67.28\%) and reduces performance from 47.70\% to 40.43\% on $\tau^2$-bench. It also trails the No Skill baseline on both benchmarks. A fixed refinement workflow is therefore not sufficient when the optimizer cannot turn execution experience into useful revisions. After WER training, the same backbone improves by 10.02 points on BFCL v4 and 12.50 points on $\tau^2$-bench. Since the backbone and refinement interface are fixed, this comparison measures the learned ability to diagnose trajectories, extract reusable lessons, and revise skills from feedback. WER has the highest average among the evaluated methods on both benchmarks, including Skill-R1, Trace2Skill, and SkillOpt.
\subsection{Phase-Wise Training Improves Iterative Refinement}
\label{sec:phase-wise-analysis}

\begin{table}[H]
    \centering
    \small
    \renewcommand{\arraystretch}{1.12}
    \begin{tabular*}{\columnwidth}{@{\extracolsep{\fill}}cccccc@{}}
        \toprule
        Phase & File Sys. & Trading & Travel & Vehicle & Avg.$\uparrow$ \\
        \midrule
        1 & 60.53 & 76.32 & 62.16 & \textbf{78.38} & 69.35 \\
        2 & \textbf{73.68} & 81.58 & 59.64 & 70.27 & 71.29 \\
        3 & 71.05 & \textbf{86.84} & \textbf{70.27} & \textbf{78.38} & \textbf{76.63} \\
        \bottomrule
    \end{tabular*}
    \caption{Phase-wise BFCL v4 performance (Pass@1, \%). Best results are in \textbf{bold}.}
    \label{tab:phase-wise-results}
\end{table}
We examine how skill-refinement performance evolves across training phases. This comparison tests whether later phases provide a distinct training benefit. We evaluate the checkpoint from each training phase under the same protocol (see \hyperref[par:evaluation-protocol]{Evaluation Protocol}).
Average Pass@1 increases monotonically from 69.35\% after Phase~1 to 71.29\% after Phase~2 and 76.63\% after Phase~3, an overall gain of 7.28 points. Individual domains fluctuate, but the final checkpoint has the best average, leads in Trading and Travel, and ties for the best Vehicle score. This pattern supports phase-wise self-bootstrapping: training on refinement states induced by earlier optimizer outputs progressively teaches the model to diagnose newly exposed failures and revise skills over multiple rounds.
\begin{figure}[!t]
    \centering
    \includegraphics[width=\columnwidth]{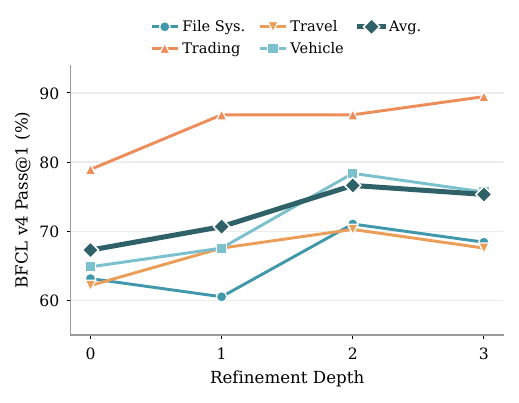}
    \caption{BFCL v4 performance trends across refinement depths.}
    \label{fig:refinement-depth}
\end{figure}

\subsection{Refinement Gains Saturate After a Few Rounds}
\label{sec:refinement-depth-analysis}

Given that phase-wise training helps, we next ask how many successive revisions are useful at inference time. Each run starts from the same GPT-5.1 seed skills. Depth~0 denotes the seeds, and each subsequent depth adds one WER revision.
\begin{table}[!htbp]
    \centering
    \small
    \renewcommand{\arraystretch}{1.12}
    \begin{tabular*}{\columnwidth}{@{\extracolsep{\fill}}cccccc@{}}
        \toprule
        Depth & File Sys. & Trading & Travel & Vehicle & Avg.$\uparrow$ \\
        \midrule
        0 & 63.16 & 78.94 & 62.16 & 64.86 & 67.28 \\
        1 & 60.53 & 86.84 & 67.57 & 67.57 & 70.67 \\
        2 & \textbf{71.05} & 86.84 & \textbf{70.27} & \textbf{78.38} & \textbf{76.63} \\
        3 & 68.42 & \textbf{89.47} & 67.57 & 75.68 & 75.33 \\
        \bottomrule
    \end{tabular*}
    \caption{BFCL v4 performance across refinement depths (Pass@1, \%). Best results are in \textbf{bold}.}
    \label{tab:refinement-depth}
\end{table}
Table~\ref{tab:refinement-depth} and Figure~\ref{fig:refinement-depth} show diminishing returns. The first two revisions account for almost all of the gain, raising average Pass@1 from 67.28\% to 76.63\%. A third revision provides no additional benefit and instead lowers the average slightly to 75.33\%. These results suggest that iterative skill refinement is most effective in its early rounds and gradually saturates as the number of revisions increases. Appendix~\ref{app:iterative-case-study} shows how successive revisions correct different execution failures.
\FloatBarrier
\subsection{Skill Refinement Is Not Subsumed by General Reasoning}
\label{sec:optimizer-backbone-analysis}

Existing skill-optimization frameworks rely predominantly on the in-context reasoning of general-purpose LLMs embedded in fixed workflows.
To test whether specialized optimizer training is necessary or if generic reasoning suffices, we evaluate DeepSeek-V4-Flash, GPT-5.5, Gemini 3.5 Flash, and Claude Sonnet 4.6 as skill optimizers under the same protocol, comparing them with the WER-trained Qwen3-4B optimizer~\cite{yang2025qwen3}.
Holding the workflow fixed isolates specialized training effects from raw base-model capability.

\begin{figure}[!htbp]
    \centering
    \includegraphics[width=\columnwidth]{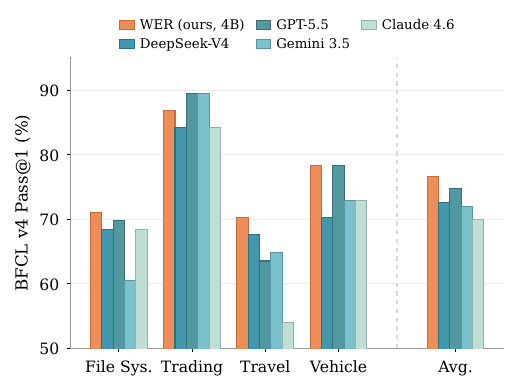}
    \caption{Skill-optimizer backbones on BFCL v4 (Pass@1, \%). The WER-trained Qwen3-4B optimizer has the highest average despite its smaller size. Per-domain results are reported in Table~\ref{tab:optimizer-backbones} (Appendix~\ref{app:optimizer-backbones}).}
    \label{fig:optimizer-backbones}
\end{figure}

Figure~\ref{fig:optimizer-backbones} shows that the general-purpose models reach average Pass@1 scores from 69.91\% to 75.28\%. The WER-trained Qwen3-4B optimizer reaches 76.63\%, 1.35 points above GPT-5.5, the strongest general-purpose model in this comparison. WER leads in File System and Travel and ties GPT-5.5 in Vehicle. These findings indicate that skill-refinement ability is not fully subsumed by general reasoning alone. Although strong general-purpose models can serve as capable skill optimizers, WER shows that targeted training offers complementary improvements in diagnosing execution traces, extracting reusable experience, and turning that experience into effective skill revisions.

This learned refinement capability can also reduce token consumption and reliance on elaborate inference-time workflows while preserving the effectiveness of skill optimization. Appendix~\ref{app:refinement-efficiency} provides the supporting analysis.
This refinement capability may also be useful for bootstrapping general-purpose language models.
\FloatBarrier

\subsection{Qualitative Analysis: Successive Revisions Target Distinct Failures}
\label{sec:iterative-case-study}

The aggregate results show that refinement helps but do not explain how skill instructions change. We trace two WER revisions of the seed skill on a representative BFCL v4 multi-turn task involving file creation and numerical aggregation, with two rollouts per stage. The seed skill leaves both procedures underspecified, and neither rollout succeeds. The first revision instructs the agent to create a missing file before writing, resolving the file-operation error but leaving one aggregation failure. The second requires collecting all relevant values before aggregation and rounding only the final result; both rollouts then succeed. The progression from 0/2 to 1/2 to 2/2 illustrates how repeated execution identifies unresolved errors and guides complementary procedural corrections. Figure~\ref{fig:iterative-case-study} in Appendix~\ref{app:iterative-case-study} provides the complete skills and trajectories.


\section{Conclusion}
\label{sec:conclusion}
WER trains a 4B skill optimizer using verified execution feedback and phase-wise self-bootstrapping while keeping the executor frozen. Under an identical refinement workflow, it improves average Pass@1 over its untrained backbone by 10.02 and 12.50 percentage points on BFCL v4 and $\tau^2$-bench, respectively.
Its 4B optimizer also surpasses all evaluated general-purpose models in this role on BFCL v4.
These results support skill refinement as a distinct learnable capability.


\section*{Limitations}
Our study has two primary limitations.
First, WER currently relies on deterministic, programmatic verifiers in stateful environments (such as BFCL v4 and $\tau^2$-bench) to provide rigorous credit assignment and select balanced mixed-outcome pairs.
While this avoids the hallucination risks of model-based evaluators, extending this paradigm to open-ended, subjective tasks lacking programmatic ground truth remains an open challenge.
Additionally, the zero-shot initialization of skills in our current pipeline leverages a frontier model, and future work should explore training the optimizer to initiate skills entirely from scratch.

Second, WER incorporates complete, verbatim interaction trajectories to construct contrastive refinement states.
Although this preserves uncompressed execution evidence crucial for diagnosing subtle tool-calling errors, it substantially increases prompt length (up to 19k tokens) as task horizons extend.
This design may incur high context-window overhead in extremely long-horizon workflows.
Investigating selective trajectory summarization or structured failure distillation without losing critical diagnostic details is a promising direction to improve context scalability.

\section*{Ethical Considerations}
Our experiments are conducted in simulated benchmark sandboxes.
While improved tool use enhances procedural automation, real-world deployments should maintain standard access controls and human oversight for sensitive actions.
Importantly, because WER represents skills as human-readable natural language documents rather than opaque model weights, the optimized instructions can be readily inspected and audited before execution.
Finally, we train a lightweight 4B model to keep the overall computational and environmental footprint modest.

\bibliography{references}

\appendix
\Needspace{32\baselineskip}

\section{Overall Training Procedure}
\label{app:training-procedure}

Algorithm~\ref{alg:training} summarizes the phase-wise training procedure described in Section~\ref{sec:method}.

\begin{algorithm}[H]
\caption{Phase-Wise Training of the Skill Optimizer}
\label{alg:training}
\small
\begin{algorithmic}[1]
\Require initial refinement states $\mathcal{X}^{(0)}$; optimizer policy $\pi_\theta$; frozen downstream executor $\pi_A$; instrumented environment $\mathcal{E}$; phases $P$; candidates per state $K$; rollouts per candidate $n$
\Ensure trained Skill Optimizer $\pi_\theta$
\For{$p=0,\ldots,P-1$}
    \State $\mathcal{B}^{(p)} \gets \varnothing$ \Comment{initialize experience buffer for phase $p$}
    \ForAll{$x=(q,\mathcal{C},h,s)\in\mathcal{X}^{(p)}$}
        \State Sample completions $o_1,\ldots,o_K\sim\pi_\theta(\cdot\mid x)$ and extract skills $s'_1,\ldots,s'_K$
        \For{$k=1,\ldots,K$}
            \For{$j=1,\ldots,n$}
                \State $\tau_{k,j}\gets\Call{Execute}{\pi_A,s'_k,\mathcal{E}}$
            \EndFor
            \State $R_k\gets\Call{Reward}{o_k, s'_k, \{\tau_{k,j}\}_{j=1}^{n};\, t}$ \Comment{Eq.~\ref{eq:reward}}
            \State Add $(x,s'_k,\{\tau_{k,j}\}_{j=1}^{n},R_k)$ to $\mathcal{B}^{(p)}$
        \EndFor
        \State Compute group advantages $\hat{A}_1,\ldots,\hat{A}_K$ from $R_1,\ldots,R_K$ \Comment{Eq.~\ref{eq:adv}}
    \EndFor
    \State Update $\theta$ on collected groups with clipped GRPO \Comment{Eq.~\ref{eq:grpo}}
    \State $\mathcal{X}^{(p+1)}\gets
    \operatorname{Assemble}\!\left(\operatorname{Retain}(\mathcal{B}^{(p)})\right)$
    \Comment{pool matched-outcome pairs}
\EndFor
\State \Return $\pi_\theta$
\end{algorithmic}
\end{algorithm}

Here, $\mathcal{X}^{(0)}$ is initialized from externally generated seed skills (instantiated by GPT-5.1) paired with their baseline execution histories.
In line~15, $\operatorname{Retain}(\mathcal{B}^{(p)})$ filters candidate records that achieve mixed execution outcomes ($n=2$, score~$1$) across the entire phase, and $\operatorname{Assemble}(\cdot)$ pairs their contrasting successful and failed trajectories to form the successor refinement states $x' = (q, \mathcal{C}, h', s')$ for phase $p+1$.

\section{Additional Experimental Details}
\label{app:additional-experimental-details}

\subsection{Benchmark Details}
\label{app:benchmark-details}

\paragraph{BFCL v4.}
We use the 200 multi-turn-base tasks from BFCL v4~\cite{patil2025berkeley}, which are evenly distributed across File System, Vehicle Control, Trading Bot, and Travel Booking. Each task may require multiple tool calls over several conversational turns, requiring the agent to maintain interaction context, use intermediate tool outputs, and execute a coherent sequence of state-dependent actions. The available tools include both domain-specific APIs and cross-functional utilities such as messaging and mathematical operations. Evaluation is execution based: the predicted calls are run in the corresponding environment, and success is determined from the resulting environment state rather than exact trajectory matching.

\paragraph{$\tau^2$-bench.}
$\tau^2$-bench~\cite{barres2025tau} evaluates conversational tool agents in the Airline, Retail, and Telecom domains. Each domain supplies a policy, task-specific tools, a stateful environment, and an LLM-based user simulator. Airline and Retail center on policy-constrained customer-service workflows, whereas Telecom introduces a dual-control setting in which the agent and user operate distinct tools over a shared environment. The agent must therefore not only reason about the task and call its own tools, but also communicate effectively and guide actions that only the user can perform. Tasks are scored by verifiable outcomes, including the final environment state and required information communicated to the user, rather than by matching a single reference trajectory. For both benchmarks, we report Pass@1 as the single-attempt task success rate.

\subsection{Baseline Details}
\label{app:baseline-details}

\paragraph{Skill-R1.}
Skill-R1~\cite{vishe2026skillr1} formulates skill optimization as a recurrent reinforcement-learning problem while keeping the task model frozen. A lightweight skill generator conditions on the task context, previous rollouts, and their verified outcomes to produce the next skill revision. Its bi-level group-relative objective combines intra-generation advantages, which compare rollouts under the same skill, with inter-generation advantages that reward improvements across successive revisions.

\paragraph{Trace2Skill.}
Trace2Skill~\cite{ni2026trace2skill} constructs transferable skills from a pool of agent execution trajectories without updating model parameters. Multiple analyst agents process trajectories in parallel to extract trajectory-local lessons and propose skill patches; these patches are then consolidated hierarchically into a unified, conflict-free skill directory. This design supports both refining an existing skill and constructing one from an initial weak draft while reducing sensitivity to any single trajectory.

\paragraph{SkillOpt.}
SkillOpt~\cite{yang2026skillopt} optimizes an external skill document through controlled text edits while keeping the task model frozen. An optimizer model analyzes successful and failed execution trajectories, proposes add/delete/replace edits under a textual update budget, and accepts a candidate skill only when it improves performance on a held-out validation split. Rejected edits are retained as negative feedback, while epoch-wise slow/meta updates consolidate reusable lessons across optimization steps.

\subsection{Training Details}
\label{app:training-details}
Optimizer training is conducted with the verl framework~\cite{sheng2025hybridflow} on a single node. We use GRPO with a batch size of 6 and sample four rollouts per prompt. The learning rate is set to $1\times10^{-6}$ following a cosine schedule with warmup. The maximum prompt and response lengths are 19,000 and 4,096 tokens, respectively, and rollouts are sampled with a temperature of 0.95 and top-$k$ sampling with $k=50$.

\paragraph{Reward Details.}
In Eq.~\ref{eq:reward}, the format weight is annealed via $\alpha_t = 1.5 - 0.5\,\operatorname{clip}((t - 1)/125,\, 0,\, 1)$ over global step $t \ge 1$ without resetting at phase boundaries.
$R_{\mathrm{len}}(o; t)$ evaluates the reasoning token length $L$: it assigns $1$ for $0 < L \le 1024$, applies a progressively stricter Gaussian penalty for $1024 < L < 2048$, and drops to $0$ for empty reasoning or $L \ge 2048$.

\subsection{Per-Domain Results for Skill-Optimizer Backbones}
\label{app:optimizer-backbones}
Table~\ref{tab:optimizer-backbones} reports the full per-domain breakdown for the skill-optimizer backbone comparison summarized in Figure~\ref{fig:optimizer-backbones}.

\begin{table}[H]
    \centering
    \small
    \setlength{\tabcolsep}{2pt}
    \renewcommand{\arraystretch}{1.18}
    \sisetup{mode=text, reset-text-series=false, text-series-to-math=true, table-format=2.2,
        table-number-alignment=center, table-model-setup=\bfseries}
    \begin{tabular}{
        >{\raggedright\arraybackslash}p{\dimexpr0.37\columnwidth-2\tabcolsep\relax}
        *{5}{S[table-column-width=\dimexpr0.126\columnwidth-2\tabcolsep\relax]}}
        \toprule
        Optimizer & {\makecell{File\\Sys.}}
        & {\makebox[0pt][c]{Trading}} & {\makebox[0pt][c]{Travel}}
        & {\makebox[0pt][c]{Vehicle}} & {\makebox[0pt][c]{Avg.$\uparrow$}} \\
        \midrule
        DeepSeek-V4-Flash & 68.42 & 84.21 & 67.56 & 70.27 & 72.61 \\
        GPT-5.5 & 69.74 & \bfseries 89.48 & 63.52 & \bfseries 78.38 & 75.28 \\
        Gemini 3.5 Flash & 60.53 & \bfseries 89.48 & 64.86 & 72.97 & 71.96 \\
        Claude Sonnet 4.6 & 68.42 & 84.21 & 54.05 & 72.97 & 69.91 \\
        \midrule
        \rowcolor{blue!8}
        \textbf{WER (Qwen3-4B)} & \bfseries 71.05 & 86.84 & \bfseries 70.27 & \bfseries 78.38 & \bfseries 76.63 \\
        \bottomrule
    \end{tabular}
    \caption{BFCL v4 performance with different skill optimizers (Pass@1, \%). Best results are in \textbf{bold}.}
    \label{tab:optimizer-backbones}
\end{table}

\subsection{Sensitivity to the Skill-Merging Model}
\label{app:merging-model-analysis}

We assess whether WER's gains depend on GPT-5.5's ability to merge skills. We hold the WER-refined skills and the GPT-4o executor fixed and vary only the model used for skill merging. This comparison isolates the effect of the merging model from that of the trained optimizer.

\begin{table}[!t]
    \centering
    \small
    \setlength{\tabcolsep}{2pt}
    \renewcommand{\arraystretch}{1.18}
    \sisetup{mode=text, reset-text-series=false, text-series-to-math=true, table-format=2.2,
        table-number-alignment=center, table-model-setup=\bfseries}
    \begin{tabular}{
        >{\raggedright\arraybackslash}p{\dimexpr0.37\columnwidth-2\tabcolsep\relax}
        *{5}{S[table-column-width=\dimexpr0.126\columnwidth-2\tabcolsep\relax]}}
        \toprule
        Merging Model & {\makecell{File\\Sys.}}
        & {\makebox[0pt][c]{Trading}} & {\makebox[0pt][c]{Travel}}
        & {\makebox[0pt][c]{Vehicle}} & {\makebox[0pt][c]{Avg.$\uparrow$}} \\
        \midrule
        DeepSeek-V4-Flash & 71.05 & \bfseries 86.84 & 67.57 & 70.27 & 73.93 \\
        GPT-4o & 63.16 & 84.21 & \bfseries 72.97 & 75.68 & 74.00 \\
        Qwen3-4B & \bfseries 73.68 & 78.95 & 67.57 & \bfseries 78.38 & 74.64 \\
        \midrule
        \rowcolor{blue!8}
        \textbf{GPT-5.5} & 71.05 & \bfseries 86.84 & 70.27 & \bfseries 78.38 & \bfseries 76.63 \\
        \bottomrule
    \end{tabular}
    \caption{BFCL v4 performance with different skill-merging models (Pass@1, \%). All settings use the WER-trained Qwen3-4B optimizer. The GPT-5.5 row reproduces the WER results in Table~\ref{tab:main-results}. Best results are in \textbf{bold}.}
    \label{tab:merging-models}
\end{table}

Table~\ref{tab:merging-models} shows that GPT-5.5 achieves the highest domain-averaged Pass@1 at 76.63\%, compared with 73.93--74.64\% for the alternative models. We therefore use GPT-5.5 as the shared merging model in the main comparisons, ensuring that evaluation uses the best-performing merger among those tested. Nevertheless, replacing GPT-5.5 changes average Pass@1 by only 1.99--2.70 percentage points, while retaining gains of 6.65--7.36 points over the GPT-5.1 seed-skill baseline. The effect of the merging model is thus modest relative to these gains, suggesting limited dependence on GPT-5.5's merging capability.

\subsection{Inference Efficiency and Refinement Workflow Analysis}
\label{app:refinement-efficiency}
\begin{figure}[!t]
    \centering
    \includegraphics[width=\columnwidth]{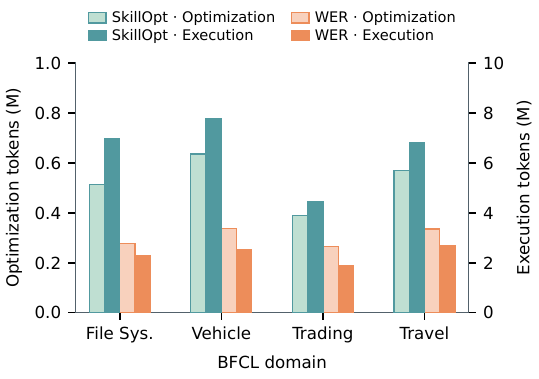}
    \caption{Token consumption of SkillOpt and WER across the four BFCL v4 domains. Optimization tokens are shown on the left axis and execution tokens on the right axis. Both axes report millions of tokens.}
    \label{fig:refinement-token-cost}
\end{figure}

Carefully orchestrated workflows can support effective skill optimization, but may incur substantial token costs. In this supplementary experiment, we decompose total token consumption into execution costs for trajectory collection and optimization costs for skill revision. We compare WER with SkillOpt~\cite{yang2026skillopt}, the strongest workflow-based baseline in our evaluation. 
Figure~\ref{fig:refinement-token-cost} shows that WER consumes fewer tokens in both stages across all four BFCL v4 domains. 
Together with the results in Table~\ref{tab:main-results}, this comparison shows that the WER-trained Skill Optimizer achieves effective skill optimization with a simpler workflow and substantially lower token consumption.

\FloatBarrier

\section{Extended Case Study Analysis}
\label{app:iterative-case-study}

\begin{figure*}[p]
    \centering
    \includegraphics[width=\textwidth]{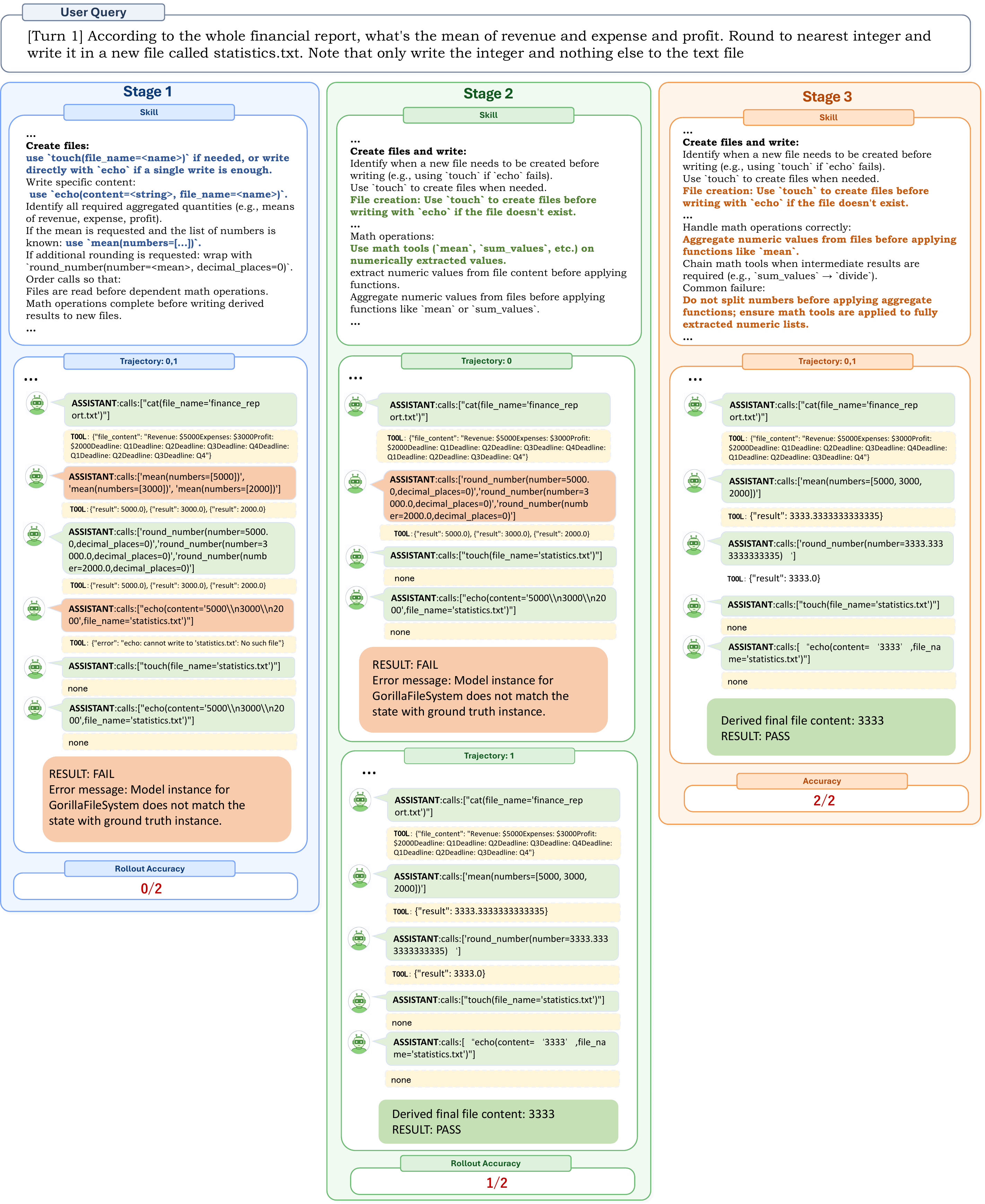}
    \caption{Three-stage skill refinement on a representative BFCL v4 multi-turn task.}
    \label{fig:iterative-case-study}
\end{figure*}

To illustrate how WER improves a skill over successive rounds, Figure~\ref{fig:iterative-case-study} traces three refinement stages on a representative BFCL v4 multi-turn task. The task requires the agent to read a financial report, aggregate several values, and write the rounded result to a newly created file. At each stage, we evaluate the current skill with two independent agent rollouts. In the visualization, colored highlights identify the key skill instructions introduced or reinforced at each stage, while orange and green result boxes denote failed and successful rollouts, respectively.

The initial seed skill leaves both file creation and numerical aggregation insufficiently specified, causing both rollouts to fail. After observing these failures, the first revision explicitly instructs the agent to create a missing file before writing to it. This eliminates the file-operation error, although one rollout still computes the aggregate incorrectly. The second revision then clarifies that all relevant values should be collected before aggregation and that rounding should be applied only to the final result. With these complementary corrections, both rollouts succeed. The progression from 0/2 to 1/2 and finally 2/2 successful rollouts provides a concrete example of WER addressing newly exposed failure modes across successive revisions.

\clearpage
\onecolumn
\section{Prompt Templates}
\label{app:prompt-templates}

\subsection{Skill Revision Prompt}
\begin{figure}[H]
    \centering
    \includegraphics[width=0.88\textwidth]{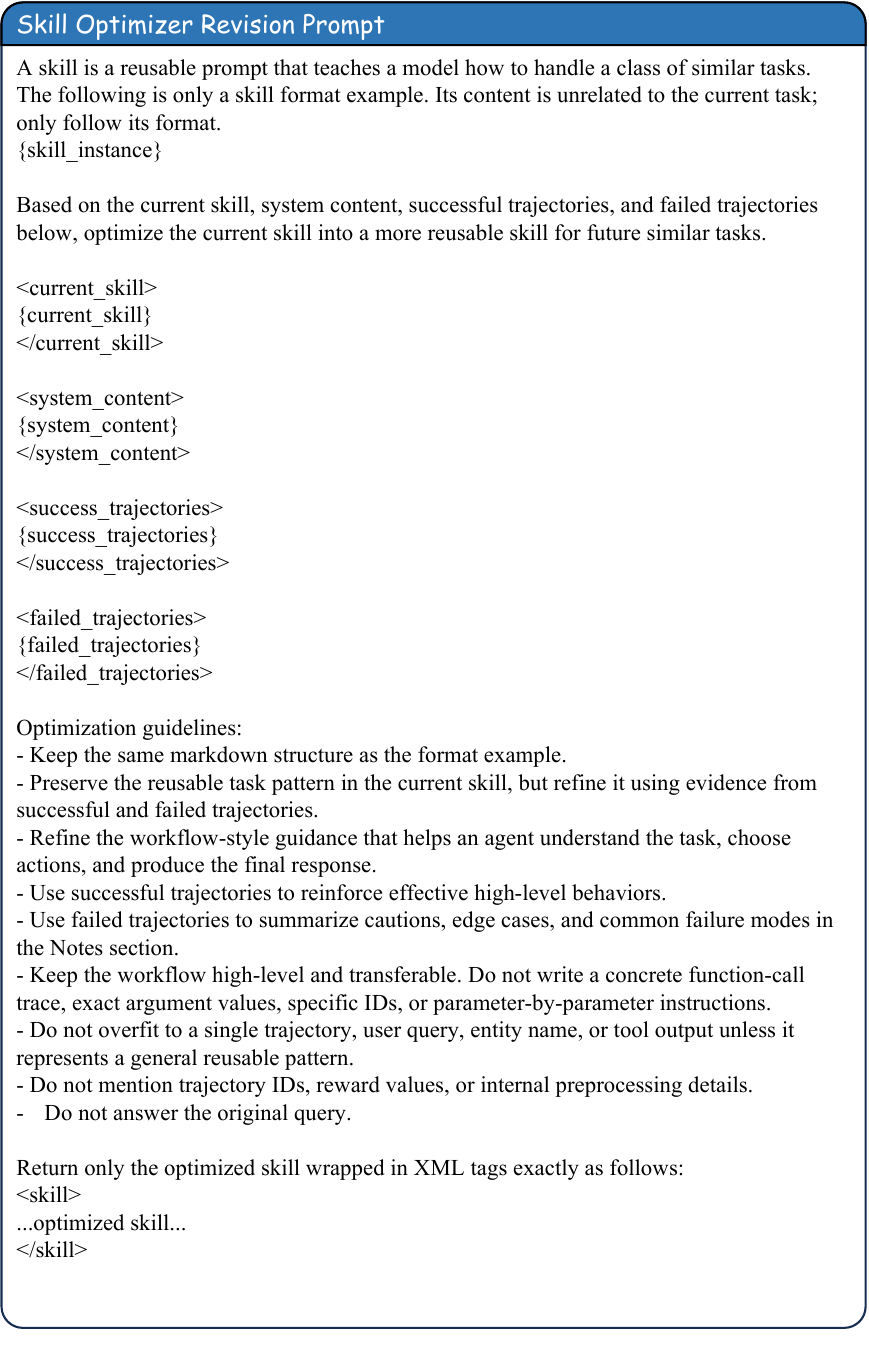}
    \caption{Prompt used by the skill optimizer to revise a skill from successful and failed execution trajectories.}
    \label{fig:skill-revision-prompt}
\end{figure}

\clearpage
\subsection{Skill Merging Prompt}
\begin{figure}[H]
    \centering
    \includegraphics[width=0.88\textwidth]{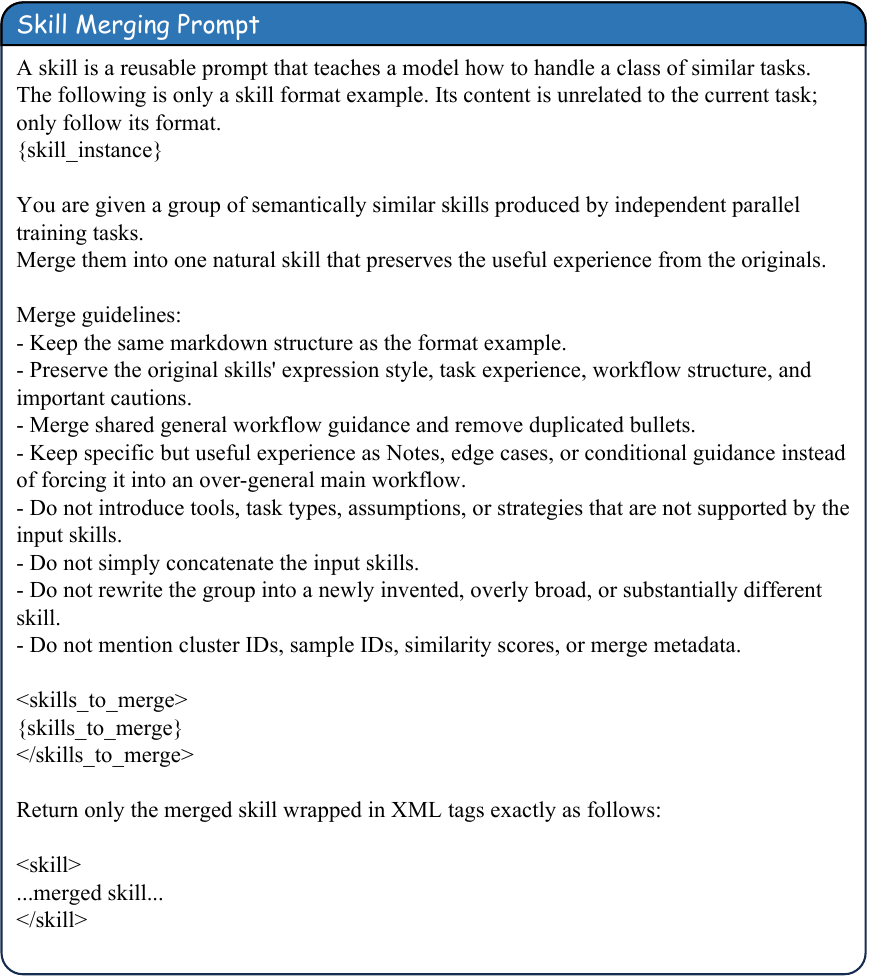}
    \caption{Prompt used to consolidate a set of related skills into a single reusable skill.}
    \label{fig:skill-merging-prompt}
\end{figure}

\end{document}